\documentclass{article}

\usepackage{microtype}
\usepackage{graphicx}
\usepackage{subfigure}
\usepackage{booktabs} 

\usepackage{amsfonts}
\usepackage{amsmath}
\usepackage{hyperref}

\usepackage[accepted]{icml2020}

\icmltitlerunning{A Generative Model for Molecular Distance Geometry}

\begin{document}

\twocolumn[
\icmltitle{A Generative Model for Molecular Distance Geometry}



\icmlsetsymbol{equal}{*}

\begin{icmlauthorlist}
\icmlauthor{Gregor N.\ C.\ Simm}{equal,cam}
\icmlauthor{Jos\'e Miguel Hern\'andez-Lobato}{cam}
\end{icmlauthorlist}

\icmlaffiliation{cam}{Department of Engineering, University of Cambridge, Cambridge, UK}

\icmlcorrespondingauthor{Gregor N.\ C.\ Simm}{gncs2@cam.ac.uk}
\icmlkeywords{graph neural networks, variational autoencoders, distance geometry, molecular conformation}

\vskip 0.3in
]



\printAffiliationsAndNotice{}  

\begin{abstract}
Great computational effort is invested in generating equilibrium states for molecular systems using, for example, Markov chain Monte Carlo.
We present a probabilistic model that generates statistically independent samples for molecules from their graph representations.
Our model learns a low-dimensional manifold that preserves the geometry of local atomic neighborhoods
through a principled learning representation that is based on Euclidean distance geometry.
In a new benchmark for molecular conformation generation, we show experimentally that our generative model achieves state-of-the-art accuracy.
Finally, we show how to use our model as a proposal distribution in an importance sampling scheme to compute molecular properties.
\end{abstract}

\section{Introduction}

\begin{figure}[t]
\center
\includegraphics{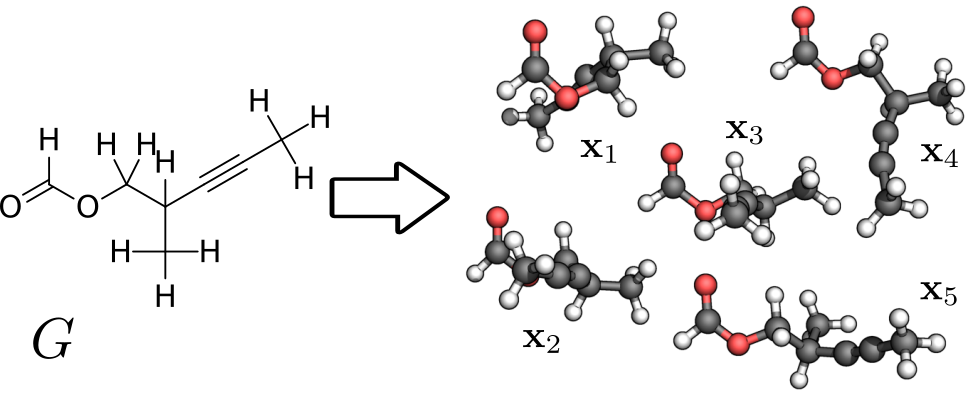}
\caption{
Standard graph representation $G$ of a molecule (left) with a set of possible conformations $\{\mathbf{x}_i\}$ (right).
It is the goal of this work to generate such conformations from the graph representation of a molecule.
Conformations feature the same atom types and bonds but the atoms are arranged differently in space.
These differences arise from rotations around and stretching of bonds in the molecule.
Hydrogen (H), carbon (C), and oxygen (O) atoms are colored white, gray, and red, respectively.
}
\label{fig:conformation}
\end{figure}

Over the last few years, many highly-effective deep learning methods generating small molecules with desired properties
(e.g., novel drugs) have emerged \citep{Gomez-Bombarelli2016,Segler2018,Dai2018,Jin2018,Bradshaw2018,Liu2018,You2018,Bradshaw2019}.
These methods operate using graph representations of molecules in which nodes and edges represent atoms and bonds, respectively.
A representation that is closer to the physical system is one in which a molecule is described by its geometry or \textit{conformation}.
A conformation $\mathbf{x}$ of a molecule is defined by a set of atoms
$\{ (\epsilon_i, \mathbf{r}_i)\}_{i=1}^{N_v}$,
where $N_v$ is the number of atoms in the molecule,
$\epsilon_i \in \{\text{H}, \text{C}, \text{O},... \}$ is the chemical element of the atom $i$,
and $\mathbf{r}_i \in \mathbb{R}^3$ is its position in Cartesian coordinates.
Importantly, the relative positions of the atoms are restricted by the bonds in the molecule and the angles between them.
Due to thermal fluctuations resulting in stretching of and rotations around bonds, there exist infinitely many conformations of a molecule.
A molecule's graph representation and a set of its conformations are shown in Fig.~\ref{fig:conformation}.
Under a wide range of conditions, the probability $p(\mathbf{x})$ of a conformation $\mathbf{x}$,
is governed by the Boltzmann distribution and is proportional to $\exp\{ - E(\mathbf{x}) / k_B T \}$,
where $E(\mathbf{x})$ is the conformation's energy,
$k_B$ is the Boltzmann constant,
and $T$ is the temperature.

To compute a molecular property for a molecule, one must sample from $p(\mathbf{x})$.
The main approach is to start with one conformation and make small changes to it over time,
e.g., by using Markov chain Monte Carlo (MCMC) or molecular dynamics (MD).
These methods can be used to accurately sample equilibrium states of molecules,
but they become computationally expensive for larger ones \citep{Shim2011,Ballard2015,DeVivo2016}.
Other heuristic approaches exist in which distances between atoms are set to fixed idealized values \citep{Havel2002,Blaney2007}.
Several methods based on statistical learning have also recently been developed to tackle the issue of conformation generation.
However, they are mainly geared towards studying proteins and their folding dynamics \citep{AlQuraishi2019}.
Some of these models are not targeting a distribution over conformations but the most stable folded configuration, e.g. \textit{AlphaFold} \citep{Senior2020},
while others are not transferable between different molecules \citep{Lemke2019,Noe2019}.

This work includes the following key contributions:
\begin{itemize}
  \item We introduce a novel probabilistic model for learning conformational distributions of molecules with graph neural networks.
  \item We create a new, challenging benchmark for conformation generation, which is made publicly available.
  To the best of our knowledge, this is the first benchmark of this kind.
  \item By combining a conditional variational autoencoder (CVAE) with an Euclidean distance geometry (EDG) algorithm
  we present a state-of-the-art approach for generating one-shot samples of molecular conformations for unseen molecules
  that is independent of their size and shape.
  \item We develop a rigorous experimental approach for evaluating and comparing the accuracy of conformation generation methods
  based on the mean maximum deviation distance metric.
  \item We show how this generative model can be used as a proposal distribution in an importance sampling (IS) scheme to estimate molecular properties.
\end{itemize}

\section{Method}

\begin{figure*}[!ht]
\centering
\includegraphics{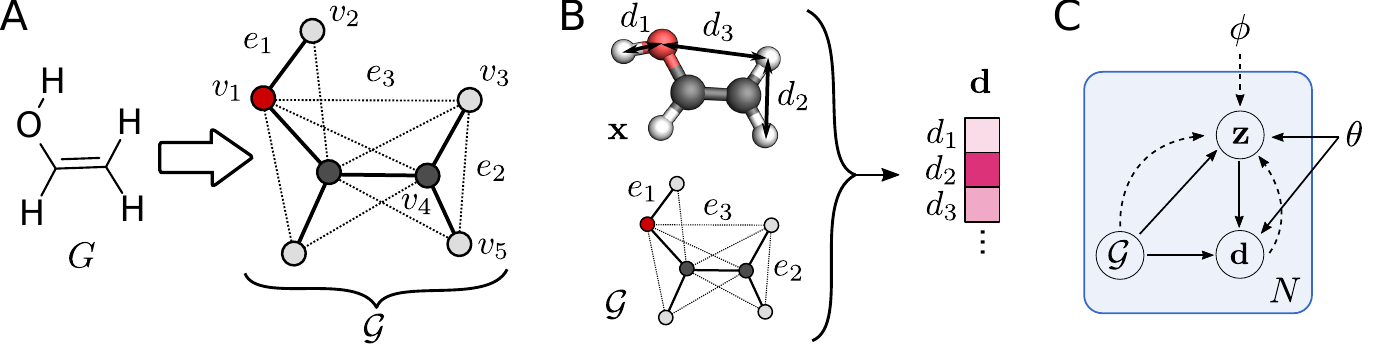}
\caption{
A) The structural formula of a molecule $G$ is converted to an extended molecular graph $\mathcal{G}$
consisting of nodes representing atoms (circles, e.g., $v_{1}$) and edges representing
molecular bonds (solid lines, e.g., $e_{1} \in E_\text{bond}$) and auxiliary edges
(dotted lines, e.g., $e_{2} \in E_\text{angle}$ and $e_{3} \in E_\text{dihedral}$).
B) The distances $\mathbf{d}$ are extracted from a conformation $\mathbf{x}$ based on the edges $E$.
C) Graphical model of the variational autoencoder: generative model
$p_\theta(\mathbf{d}| \mathbf{z}, \mathcal{G}) p_\theta(\mathbf{z}| \mathcal{G})$ (solid lines)
and variational approximation $q_\phi(\mathbf{z} | \mathbf{d}, \mathcal{G})$ (dashed lines).
}
\label{fig:method}
\end{figure*}

Our goal is to build a statistical model that generates molecular conformations in a one-shot fashion from a molecule's graph representation.
First, we describe how a molecule's conformation can be represented by a set of pairwise distances between atoms
and why this presentation is advantageous over one in Cartesian coordinates (Section~\ref{sec:graphs}).
Second, we present a generative model in Section~\ref{sec:model} that will generate sets of atomic distances for a given molecular graph.
Third, we explain in Section~\ref{sec:dg} how a set of predicted distances can be transformed into a molecular conformation
and why this transformation is necessary.
Finally, we detail in Section~\ref{sec:is} how our generative model can be used as a proposal distribution
in an IS scheme to estimate molecular properties.

\subsection{Extended Molecular Graphs and Distance Geometry}
\label{sec:graphs}

In this study, a molecule is represented by an undirected graph which is defined as a tuple $G = (V, E)$.
$V = \{v_i\}_{i=1}^{N_v}$ is the set of nodes representing atoms, where each $v_i \in \mathbb{R}^{F_v}$ holds atomic attributes (e.g., the element type $\epsilon_i$).
$E = \{(e_k , r_k , s_k)\}_{k=1}^{N_e}$ is the set of edges, where each $e_k \in \mathbb{R}^{F_e}$ holds an edge's attributes (e.g., the bond type),
and $r_k$ and $s_k$ are the nodes an edge is connecting.
Here, $E$ represents the molecular bonds in the molecule.

We assume that, given a molecular graph $G$, one can represent one of its conformations $\mathbf{x}$ by a set of atomic distances $\mathbf{d} = \{d_{k}\}_{k=1}^{N_e}$,
where $d_{k} = | \mathbf{r}_{r_k} - \mathbf{r}_{s_k} |$ is the Euclidean distance between the positions of the atoms $r_k$ and $s_k$ in this conformation.
As the set of edges between the bonded atoms ($E_\text{bond}$) alone would not suffice to describe a conformation,
we expand the traditional graph representation of a molecule by adding \textit{auxiliary} edges to obtain an extended graph $\mathcal{G}$.
Auxiliary edges between atoms that are second neighbors in the original graph $G$ fix angles between atoms,
and those between third neighbors fix dihedral angles (denoted $E_\text{angle}$ and $E_\text{dihedral}$, respectively).
In this work, $E_\text{angle}$ are added between nodes in $\mathcal{G}$ which are second neighbors in $G$.
After all $E_\text{angle}$ have been added, additional edges are added to $\mathcal{G}$ from a node
$v$ to a randomly chosen third neighbor of $v$ in $G$ if $v$ has less then three neighbors in $\mathcal{G}$.
Therefore, a graph $G$ can give rise to multiple different extend graphs $\mathcal{G}$.
In Fig.~\ref{fig:method}, the process of extending the molecular graph
and the extraction of $\mathbf{d}$ from $\mathbf{x}$ and $\mathcal{G}$ are illustrated.

A key advantage of a representation in terms of distances is its invariance to rotation and translation;
by contrast, Cartesian coordinates depend on the (arbitrary) choice of origin, for example.
In addition, it reflects pair-wise physical interactions and their generally local nature.
Auxiliary edges can be placed between higher-order neighbors depending on how far the physical interactions
dominating the potential energy of the system reach.

We have a set of $N_G$ pairs, $\{G_i, x_i\}_{i=1}^{N_G}$, consisting of a molecular graph and a conformation.
With the protocol described above, we convert each pair into a pair of an extended molecular graph together with a set of distances $\mathbf{d}$
to obtain $\{\mathcal{G}_i, \mathbf{d}_i\}_{i=1}^{N_G}$.
With this data, we will train a generative model which we detail in the following section.

\subsection{Generative Model}
\label{sec:model}

We employ a CVAE \citep{Kingma2013,Pagnoni2018} to model the distribution over distances $\mathbf{d}$ given a molecular graph $\mathcal{G}$.
A CVAE first encodes $\mathcal{G}$ together with $\mathbf{d}$ into a latent space
$\mathbf{z} \in \mathbb{R}^{k N_v}$, where $k \in \mathbb{N}^+$, with an encoder $q_\phi(\mathbf{z} | \mathbf{d}, \mathcal{G})$.
Subsequently, the decoder $p_\theta(\mathbf{d}| \mathbf{z}, \mathcal{G})$ decodes $\mathbf{z}$ back into a set of distances.
A graphical model is shown in Fig.~\ref{fig:method}, C.

A conformation has, in general, $3N_v - 6$ spatial degrees of freedom (dofs):
one dof per spacial dimension per atom minus three translational and three rotational dofs.
Therefore, the latent space should be proportional to the number of atoms in the molecule.
In addition, the latent space should be smaller than $3N_v$ as it is the role of the encoder to project the conformation into a lower-dimensional space.
As a result, we set $k=1$.%
\footnote{
Experiments showed that our model performs similarly well with a latent space of $\mathbb{R}^{2N_v}$ and $\mathbb{R}^{3N_v}$.
We chose to use $k=1$ for simplicity.
}

Here, $q_\phi(\mathbf{z} | \mathbf{d}, \mathcal{G})$ and $p_\theta(\mathbf{d} | \mathbf{z}, \mathcal{G})$ are Gaussian distributions,
the mean and variance of which are modeled by two artificial neural networks.
At the center of this model are message-passing neural networks (MPNNs) \citep{Gilmer2017}.
In short, an MPNN is a convolutional neural network that allows end-to-end learning of prediction pipelines whose inputs
are graphs of arbitrary size and shape.
In a convolution, neighboring nodes exchange so-called messages between neighbors to update their attributes.
Edges update their attributes with the features of the nodes they are connecting.
The MPNN is a well-studied technique that achieves state-of-the-art performance in representation learning for molecules
\citep{Kipf2016,Duvenaud2015,Kearnes2016,Schutt2017a,Gilmer2017,Kusner2017,Bradshaw2018}.

\begin{figure*}[t]
\centering
\includegraphics{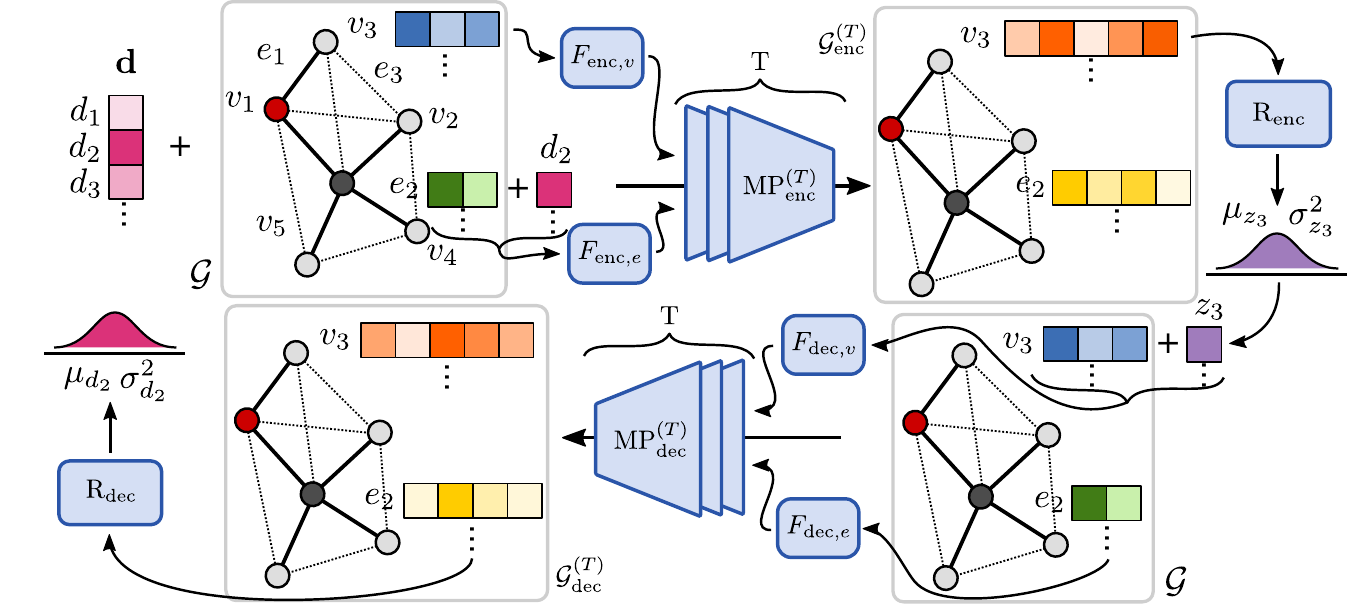}
\caption{
The molecular graph $\mathcal{G}$ together with the distances $\mathbf{d}$ are passed through the model consisting of
an encoder $q_\phi(\mathbf{z} | \mathbf{d}, \mathcal{G})$ and a decoder $p_\theta(\mathbf{d}| \mathbf{z}, \mathcal{G})$.
See Section~\ref{sec:model} for details.
}
\label{fig:model}
\end{figure*}

In the following, we describe the details of the mode which is illustrated in Fig.~\ref{fig:model}.%
\footnote{The model is available online \url{https://github.com/gncs/graphdg}}
In the encoder $q_\phi(\mathbf{z} | \mathbf{d}, \mathcal{G})$,
each $d_k$ is concatenated with the respective edge feature $e_k$ to give $e_k' \in \mathbb{R}^{F_e + 1}$.
Then, each $v_i$ and each $e'_k$ are passed to $F_{\text{enc}, v}$ and $F_{\text{enc}, e}$ (two multilayer perceptrons, MLPs), respectively,
to give $\mathcal{G}_\text{enc}^{(0)}$, where
$\mathcal{G}_\text{enc}^{(t)} = (\{v_{i,\text{enc}}^{(t)}\}_{i=1}^{N_v}, \{(e_{k,\text{enc}}^{(t)} , r_k , s_k)\}_{k=1}^{N_e})$,
$v_{i, \text{enc}}^{(t)} \in \mathbb{R}^{L_\text{v}}$, and $e_{k, \text{enc}}^{(t)} \in \mathbb{R}^{L_\text{e}}$.
Then, $T$ MPNNs of depth 1, $\{ \text{MP}^{(t)}_\text{enc} \}_{t=1}^T$, are consecutively applied to obtain $\mathcal{G}_\text{enc}^{(T)}$.
Finally, the read-out function $R_\text{enc}$ (an MLP) takes each $v_\text{i, enc}^{(T)}$ to predict the mean $\mu_{z_i} \in \mathbb{R}$ and the variance $\sigma^2_{z_i} \in \mathbb{R}$
of the Gaussian distribution for $z_i$.
The so-called reparametrization trick is employed to draw a sample for $z_i$.
In summary,
\begin{equation}
v_{i,\text{enc}}^{(0)} = F_{\text{enc}, v}(v_i), \quad
e_{k,\text{enc}}^{(0)} = F_{\text{enc}, e}(e'_i),
\end{equation}
\begin{equation}
\mathcal{G}_\text{enc}^{(t+1)} = \text{MP}^{(t)}_\text{enc} (\mathcal{G}_\text{enc}^{(t)}),
\end{equation}
\begin{equation}
\mu_{z_i}, \sigma^2_{z_i} = R_\text{enc}(v_{i,\text{enc}}^{(T)}).
\end{equation}
In the decoder $p_\theta(\mathbf{d}| \mathbf{z}, \mathcal{G})$, each $z_i$ is concatenated with the respective node feature $v_i$ to give $v_i' \in \mathbb{R}^{F_v + 1}$.
Each $v'_i$ and each $e_k$ are passed to $F_{\text{dec}, v}$ and $F_{\text{dec}, e}$ (two MLPs), respectively,
to give $\mathcal{G}_\text{dec}^{(0)}$, where
$\mathcal{G}_\text{dec}^{(t)} = (\{v_{i,\text{dec}}^{(t)}\}_{i=1}^{N_v}, \{(e_{k,\text{dec}}^{(t)} , r_k , s_k)\}_{k=1}^{N_e})$,
$v_{i, \text{dec}}^{(t)} \in \mathbb{R}^{L_\text{v}}$, and $e_{k, \text{dec}}^{(t)} \in \mathbb{R}^{L_\text{e}}$.
Then, $T$ MPNNs of depth 1, $\{ \text{MP}^{(t)}_\text{dec} \}_{t=1}^T$, are consecutively applied to obtain $\mathcal{G}_\text{dec}^{(T)}$.
Finally, the read-out function $R_\text{dec}$ (an MLP) takes each $e_\text{k, dec}^{(T)}$ to predict the mean $\mu_{d_k} \in \mathbb{R}$ and the variance $\sigma^2_{d_k} \in \mathbb{R}$
of the Gaussian distribution for $d_k$.
In summary,
\begin{equation}
v_{i,\text{dec}}^{(0)} = F_{\text{dec}, v}(v'_i), \quad
e_{k,\text{dec}}^{(0)} = F_{\text{dec}, e}(e_i), \\
\end{equation}
\begin{equation}
\mathcal{G}_\text{dec}^{(t+1)} = \text{MP}^{(t)}_\text{dec} (\mathcal{G}_\text{dec}^{(t)}),
\end{equation}
\begin{equation}
\mu_{d_k}, \sigma^2_{d_k} = R_\text{dec}(e_{k,\text{dec}}^{(T)}).
\end{equation}
The sets of parameters in the encoder and decoder, $\phi$ and $\theta$
(i.e., parameters in $F_{\text{enc}, v}$, $F_{\text{enc}, e}$, $\{ \text{MP}^{(t)}_\text{enc} \}_{t=1}^T$, $R_\text{enc}$,
$F_{\text{dec}, v}$, $F_{\text{dec}, e}$, $\{ \text{MP}^{(t)}_\text{dec} \}_{t=1}^T$, $R_\text{dec}$),
respectively, are optimized by maximizing the evidence lower bound (ELBO):
\begin{equation}
\begin{split}
L = \mathbb{E}_{\mathbf{z} \sim q_\phi(\mathbf{z} | \mathbf{d}, \mathcal{G})} [\log p_\theta(\mathbf{d} | \mathbf{z}, \mathcal{G})] \\
- D_\text{KL} [q_\phi(\mathbf{z}|\mathbf{d}, \mathcal{G}) || p_\theta(\mathbf{z} | \mathcal{G})],
\end{split}
\end{equation}
where the prior $p_\theta(\mathbf{z} | \mathcal{G})$ consists of factorized standard Gaussians.
The optimal values for the hyperparameters for the network dimensions, number of message passes, batch size, and learning rate
of the Adam optimizer \citep{Kingma2014}
were manually tuned by maximizing the validation performance (ELBO) and are reported in the Appendix.

\subsection{Conformation Generation through Euclidean Distance Geometry}
\label{sec:dg}

To compute molecular properties, quantum-chemical methods need to be employed which require the input, i.e., the molecule,
to be in Cartesian coordinates.%
\footnote{
Even though quantum-chemical methods require the input to be in Cartesian coordinates,
calculated properties, such as the energy, are invariant under translation and rotation.
}
Therefore, we use an EDG algorithm to translate the set of distances $\{ d_k \}_{k=1}^{N_e}$
to a set of atomic coordinates $\{\mathbf{r}_i\}_{i=1}^{N_v}$.%
\footnote{
There are additional constraints due to chirality.
However, since they are given by $\mathcal{G}$ and are fixed, they are not modeled by our method.
}

EDG is the mathematical basis for a geometric theory of molecular conformation.
In the field of machine learning, \citet{Weinberger2006} used it for learning image manifolds,
\citet{Tenenbaum2000} for image understanding and handwriting recognition,
\citet{Jain2004} for speech and music, and \citet{Demaine2009} for music and musical rhythms.
An EDG description of a molecular system consists of a list of lower and upper bounds on the distances between pairs of
atoms $\{ (d_{k, \text{min}}, d_{k, \text{max}}) \}_{k=1}^{N_e}$.
Here, $p_\theta(\mathbf{d} | \mathbf{z}, \mathcal{G})$ is used to model these bounds, namely, we set the bounds to
$\{ (\mu_{d_{k}} - \sigma_{d_{k}}, \mu_{d_{k}} + \sigma_{d_{k}}) \}$,
where $\mu_{d_k}$ and $\sigma_{d_k}$ are the mean and standard deviation for each distance $d_k$ given by the CVAE.
Then, an EDG algorithm determines a set of Cartesian coordinates $\{ \mathbf{r}_i \}_{i=1}^{N_v}$ so that these bounds are fulfilled
(see the Appendix for details).%
\footnote{
Often there exist multiple solutions for the same set of bounds.
As the bounds are generally tight, the solutions are very similar.
Therefore, we only generate one set of coordinates per set of bounds.
}
Together with the corresponding chemical elements $\{\epsilon_i \}_{i=1}^{N_v}$, we obtain a conformation $\mathbf{x}$.

\subsection{Calculation of Molecular Properties}
\label{sec:is}

We can get an MC estimate of the expectation $\mathbb{E}[\mathcal{O}]$ of a property $\mathcal{O}$ (e.g., the dipole moment)
for a molecule represented by $G$ by generating an extended graph $\mathcal{G}$,
drawing conformational samples $\mathbf{x}_i \sim p(\mathbf{x} | \mathcal{G})$,
and computing $\mathcal{O}(\mathbf{x}_i) \in \mathbb{R}$ with a quantum-chemical method (e.g., density functional theory).
Since we cannot draw samples from $p(\mathbf{x} | \mathcal{G})$ directly, we employ an IS integration scheme \citep{Bishop2009}
with our CVAE as the proposal distribution.
We assume that we can readily evaluate the unnormalized probability of a conformation
$\tilde{p}(\mathbf{x} | \mathcal{G}) = \exp\{ - E(\mathbf{x}) / k_B T \}$, where $\mathbf{x}$ must be a conformation of the molecule
and the energy $E(\mathbf{x})$ is determined with a quantum-chemical method.
Since the EDG algorithm is mapping the distribution $p_\theta(\mathbf{d} | \mathbf{z}, \mathcal{G})$ to a point mass in $\mathbb{R}^{3N_v}$,
the MC estimate for the resulting distribution $p_\text{prop}(\mathbf{x}| \mathcal{G})$ is approximated by a mixture of delta functions,
each of which is centered at the $\mathbf{x}_i$ resulting from mapping $p_\theta(\mathbf{d} | \mathbf{z}_i, \mathcal{G})$
to $\mathbb{R}^{3N_v}$, where $\mathbf{z}_i \sim p_\theta(\mathbf{z} | \mathcal{G})$, that is,
$p_\text{prop}(\mathbf{x} | \mathcal{G}) \approx \frac{1}{N} \sum_{i=1}^{N} \delta(\mathbf{x} - \mathbf{x}_i)$.
The IS estimator for the expectation of $\mathcal{O}$ w.\ r.\ t.\ $\tilde{p}(\mathbf{x} | \mathcal{G})$ then reads
\begin{equation}
  \mathbb{\hat{E}}_\mathcal{G}[\mathcal{O}]
  \stackrel{\text{MC}}{\approx} \frac{1}{N} \sum_{i=1}^{N} \mathcal{O}(\mathbf{x}_i)
  \stackrel{\text{IS}}{=}\frac{1}{N} \sum_{i=1}^{N} \mathcal{O}(\mathbf{x}'_i) \frac{\tilde{p}(\mathbf{x}'_i | \mathcal{G})}{p_\text{prop}(\mathbf{x}'_i | \mathcal{G})},
\end{equation}
where $\mathbf{x}_i \sim \tilde{p}(\mathbf{x}_i | \mathcal{G})$ and $\mathbf{x}'_i \sim p_\text{prop}(\mathbf{x}'_i | \mathcal{G})$,
so that the expectation of $\mathcal{O}$ w.\ r.\ t.\ the normalized version of $\tilde{p}(\mathbf{x})$ is then
\begin{equation}
  \mathbb{E}_\mathcal{G}[\mathcal{O}]
  = \frac{\mathbb{\hat{E}}_\mathcal{G}[\mathcal{O}]}{\mathbb{\hat{E}}_\mathcal{G}[1]}
  \approx \frac{1}{Z} \sum_{i=1}^{N} \mathcal{O}(\mathbf{x}_i) \tilde{p}(\mathbf{x}'_i | \mathcal{G}),
\end{equation}
where $\mathbb{\hat{E}}_\mathcal{G}[1]$ is the expectation of an operator that returns $1$ for every conformation $\mathbf{x}$,
$Z \approx \sum_{i=1}^{N} \tilde{p}(\mathbf{x}'_i \mid \mathcal{G})$,
and $N$ is the number of samples.
When dividing two delta functions we have assumed that they take some arbitrarily large finite value.

\section{Related Works}

The standard approach for generating molecular conformations is to start with one,
and make small changes to it over time, e.g., by using MCMC or MD.
These methods are considered the gold standard for sampling equilibrium states,
but they are computationally expensive, especially if the molecule is large and the Hamiltonian is based on quantum-mechanical principles \citep{Shim2011,Ballard2015,DeVivo2016}.

A much faster but more approximate approach for conformation generation is EDG \citep{Havel2002,Blaney2007,Lagorce2009,Riniker2015}.
Lower and upper distance bounds for pairs of atoms in a molecule are fixed values based on ideal bond lengths,
bond angles, and torsional angles.
These values are often extracted from crystal structure databases \citep{Allen2002}.
These methods aim to produce a low-energy conformation,
not to generate unbiased samples from the underlying distribution at a certain temperature.

There exist several machine learning approaches as well, however, they are mostly tailored towards studying protein dynamics.
For example, \citet{Noe2019} trained Boltzmann generators on the energy function of proteins to provide unbiased,
one-shot samples from their equilibrium states.
This is achieved by training an invertible neural network to learn a coordinate transformation from a system’s configurations to a latent space representation.
Further, \citet{Lemke2019} proposed a dimensionality reduction algorithm that is based on a neural network autoencoder in combination with a nonlinear distance metric
to generate samples for protein structures.
Both models learn protein-specific coordinate transformations that cannot be transferred to other molecules.

\citet{AlQuraishi2019} introduced an end-to-end differentiable recurrent geometric network for protein structure learning based on amino acid sequences.
Also, \citet{Ingraham2019} proposed a neural energy simulator model for protein structure that makes use of protein sequence information.
Recently, \citet{Senior2020} significantly advanced the field of protein-structure prediction with a new model called \textit{AlphaFold}.
In contrast to amino acid sequences, molecular graphs are, in general,
not linear but highly branched and often contain cycles.
This makes these approaches unsuitable for general molecules.

Finally, \citet{Mansimov2019} presented a conditional deep generative graph neural network to generate molecular conformations given a molecular graph.
Their goal is to predict the most likely conformation and not a distribution over conformations.
Instead of encoding molecular environments in atomic distances, they work directly in Cartesian coordinates.
As a result, the generated conformations showed significant structural differences compared to the ground-truth
and required refinement through a force field, which is often employed in MD simulations.

We argue that our model has several advantages over the approaches reviewed above:
\begin{itemize}
  \item It is a fast alternative to resource-intensive approaches based on MCMC or MD.
  \item Our principled representation based on pair-wise distances does not restrict our approach to any particular molecular structure.
  \item Our model is, in principle, transferable to unseen molecules.
\end{itemize}

\section{The \textsc{Conf17} Benchmark}
\label{sec:benchmark}

The \textsc{Conf17} benchmark is the first benchmark for molecular conformation sampling.%
\footnote{
Datasets such as the one published by \citet{Kanal2018} only include \textit{conformers}, i.e., the stable conformations of a molecule, and not a distribution over conformations.
}
It is based on the \textsc{ISO17} dataset \citep{Schutt2017}
which consists of conformations of various molecules with the atomic composition C$_7$H$_{10}$O$_2$ drawn from the \textsc{QM9} dataset \citep{Ramakrishnan2014}.
These conformations were generated by \textit{ab initio} molecular dynamics simulations at 500~Kelvin.
From the \textsc{ISO17} dataset, 430692 valid molecular graph-conformation pairs could be extracted and
197 unique molecular graphs could be identified.
We split the dataset into training and test sets such that no molecular graph in the training set can be found in the test or vice versa.
Training and test splits consist of 176 and 30 unique molecular graphs, respectively (see Appendix~\ref{sec:conf17_appendix} for details).

\begin{figure}[t]
\centering
\includegraphics{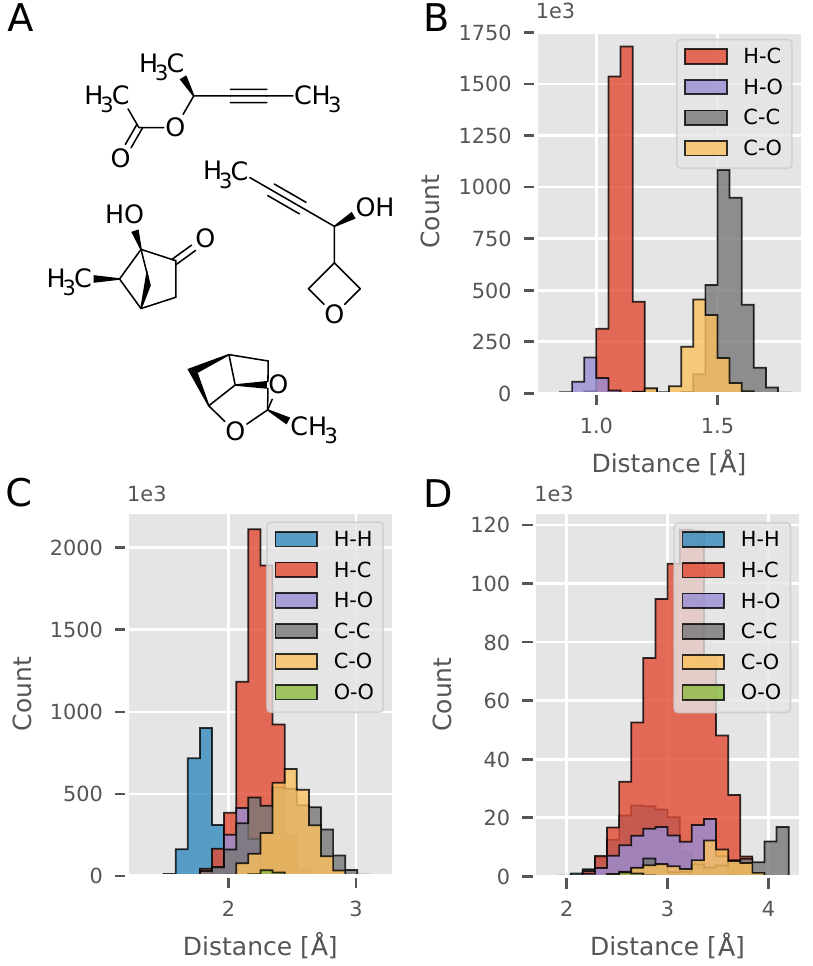}
\caption{
Overview of the \textsc{Conf17} benchmark.
A: Structural formulae of a random selection of molecules.
B--D: Distribution of distances (in \AA{}) grouped by edge (B: $E_\text{bond}$, C: $E_\text{angle}$, and D: $E_\text{dihedral}$)
and vertex type (chemical element).
}
\label{fig:dataset}
\end{figure}

In Fig.~\ref{fig:dataset}, A, the structural formulae of a random selection of molecules from this benchmark are shown.
Most molecules feature highly-strained, complex 3D structures such as rings which are typical of drug-like molecules.
It is thus the structural complexity of the molecules, not their number of degrees of freedom, that makes this benchmark challenging.
In Fig.~\ref{fig:dataset}, B--D, the frequency of distances (in \AA{}) in the conformations are shown for each edge type.
It can be seen that the marginal distributions of the edge distances are multimodal and highly context-dependent.

\section{Experiments}
\label{sec:experiments}

We assess the performance of our method, named Graph Distance Geometry (\textsc{GraphDG}), by comparing it with two state-of-the-art methods for molecular conformation generation:
\textsc{RDKit} \citep{Riniker2015}, a classical EDG approach,
and \textsc{DL4Chem} \citep{Mansimov2019}, a machine learning approach.
We trained \textsc{GraphDG} and \textsc{DL4Chem} on three different training and test splits of the \textsc{Conf17} benchmark
using Adam \citep{Kingma2014}.
We generated $100$ conformations with each method for molecular graphs in a test set.

\subsection{Distributions Over Distances}

We assessed the accuracy of the distance distributions of \textsc{RDKit}, \textsc{DL4Chem}, and \textsc{GraphDG}
by calculating the maximum mean discrepancy (MMD) \citep{Gretton2012} to the ground-truth distribution.
In particular, we compute the MMD using a Gaussian kernel, where we set the standard deviation to be the median distance between distances $\mathbf{d}$ in the aggregate sample.
For this, we determined the distances in the conformations from the ground-truth and those generated by \textsc{RDKit}, \textsc{DL4Chem}, and \textsc{GraphDG}.
For each train-test split and each $G$ in a test set,
we compute the MMD of the joint distribution of distances between C and O atoms $p(\{ d_{k} \} | G)$ (H atoms are usually ignored),
the MMDs of pair-wise distances $p(d_{i}, d_{j} | G)$,
and the MMDs between the marginals of individual distances $p(d_i | G)$.
We aggregate the results of three train-test splits,
and, finally, compute the median MMDs and average rankings.
The results are summarized in Table~\ref{tab:mmds}.
It can be seen that the samples from \textsc{GraphDG} are significantly closer to the ground-truth distribution than the other methods.
\textsc{RDKit} is slightly worse than \textsc{GraphDG} while \textsc{DL4Chem} seems to struggle with the complexity of the molecules
and the small number of graphs in the training set.

\begin{table*}
\centering
\caption{
Assessment of the accuracy of the distributions over conformations generated by three models compared to the ground-truth.
We compare the distributions with respect to the marginals $p(d_{k} | G)$,
$p(d_{k}, d_{l} | G)$,
and the distribution over all edges between C and O atoms $p(\{ d_{k} \} | G)$.
Two different metrics are used:
median MMD between ground-truth conformations and generated ones,
and mean ranking (1 to 3) based on the MMD.
Reported are the results for molecular graphs in a test set from three train-test splits.
Standard deviations are given in brackets.
}
\label{tab:mmds}
\begin{tabular}{l|rrr|rrr}
\toprule
 & \multicolumn{3}{c|}{Median MMD} & \multicolumn{3}{c}{Mean Ranking}  \\
 & \multicolumn{1}{c}{\textsc{RDKit}} & \multicolumn{1}{c}{\textsc{DL4Chem}} & \multicolumn{1}{c|}{\textsc{GraphDG}} & \multicolumn{1}{c}{\textsc{RDKit}} & \multicolumn{1}{c}{\textsc{DL4Chem}} & \multicolumn{1}{c}{\textsc{GraphDG}} \\
\midrule
$p(d_{k} | G)$         & $0.37 \; (0.23)$   &   $1.11 \; (0.25)$   &   $\mathbf{0.13} \; (0.13)$ &   $ 1.98 \; (0.44)$   &   $ 2.90 \; (0.35)$   &   $\textbf{1.12} \; (0.33)$ \\
$p(d_{k}, d_{l} | G)$  & $0.47 \; (0.18)$   &   $1.12 \; (0.15)$   &   $\mathbf{0.14} \; (0.11)$ &   $ 1.95 \; (0.29)$   &   $ 2.98 \; (0.13)$   &   $\textbf{1.07} \; (0.26)$ \\
$p(\{ d_{k} \} | G)$   & $0.57 \; (0.11)$   &   $1.03 \; (0.13)$   &   $\mathbf{0.19} \; (0.08)$ &   $ 2.00 \; (0.00)$   &   $ 3.00 \; (0.00)$   &   $\textbf{1.00} \; (0.00)$ \\
\bottomrule
\end{tabular}
\end{table*}

\begin{figure*}[t]
\begin{center}
\includegraphics{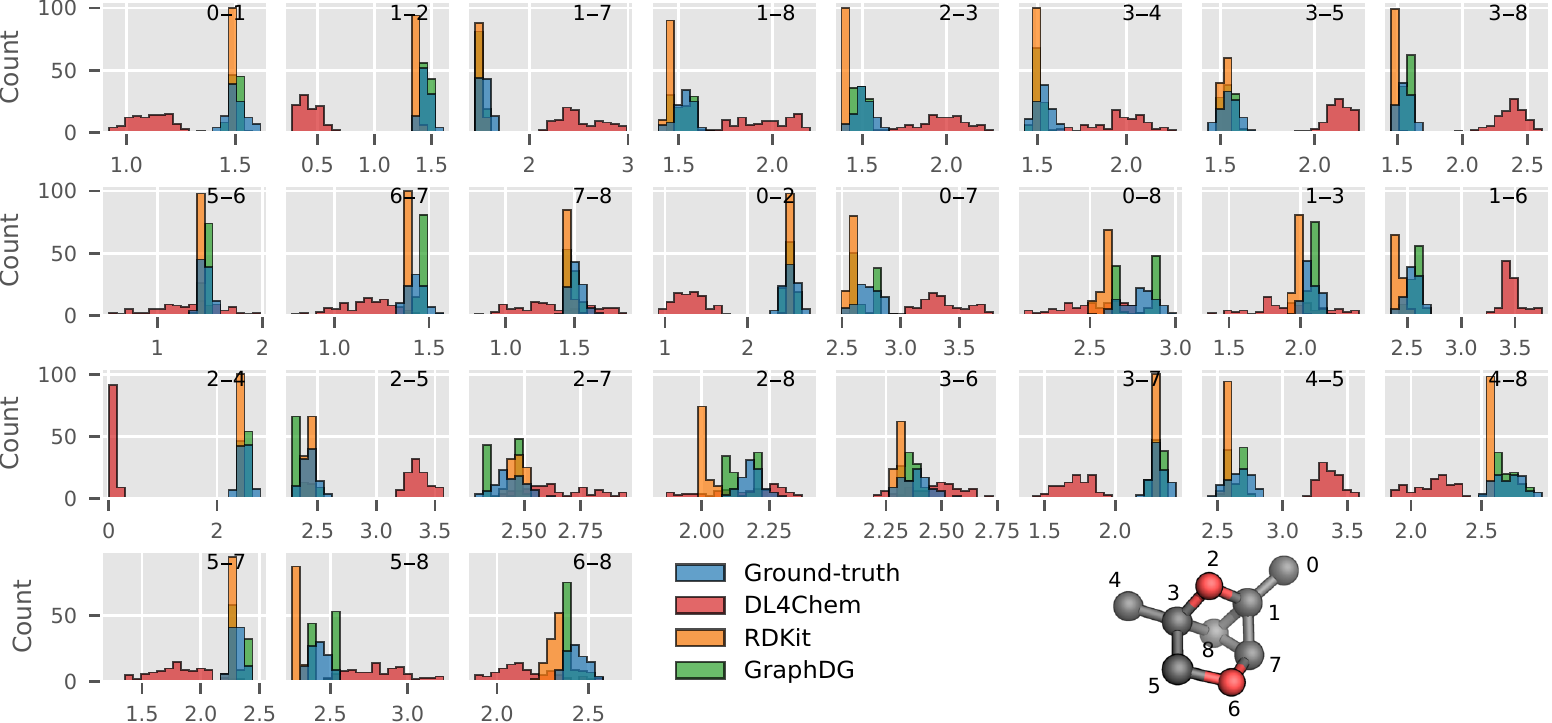}
\caption{
Marginal distributions $p(d_{k} | G)$ of ground-truth and predicted distances (in \AA{}) between C and O atoms
given a molecular graph from a test set.
The atoms connected by each edge $d_k$ are indicated in each subplot ($s_k$--$r_k$).
In the 3D structure of the molecule, carbon and oxygen atoms are colored gray and red, respectively.
H atoms are omitted for clarity.
}
\label{fig:hists}
\end{center}
\end{figure*}

In Fig.~\ref{fig:hists}, we showcase the accuracy of our model by plotting the marginal distributions
$p(d_{i} | G)$ for distances between C and O atoms, given a molecular graph from a test set.
It can be seen that \textsc{RDKit} consistently underestimates the marginal variances.
This is because this method aims to predict the most stable conformation, i.e., the distribution's mode.
In contrast, \textsc{DL4Chem} often fails to predict the correct mean.
For this molecule, \textsc{GraphDG} is the most accurate, predicting the right mean and variance in most cases.
Additional figures can be found in the Appendix,
where we also show plots for the marginal distributions $p(d_{i}, d_{j} | G)$.

\subsection{Generation of Conformations}
\begin{figure*}[!ht]
\centering
\includegraphics[width=0.9\textwidth]{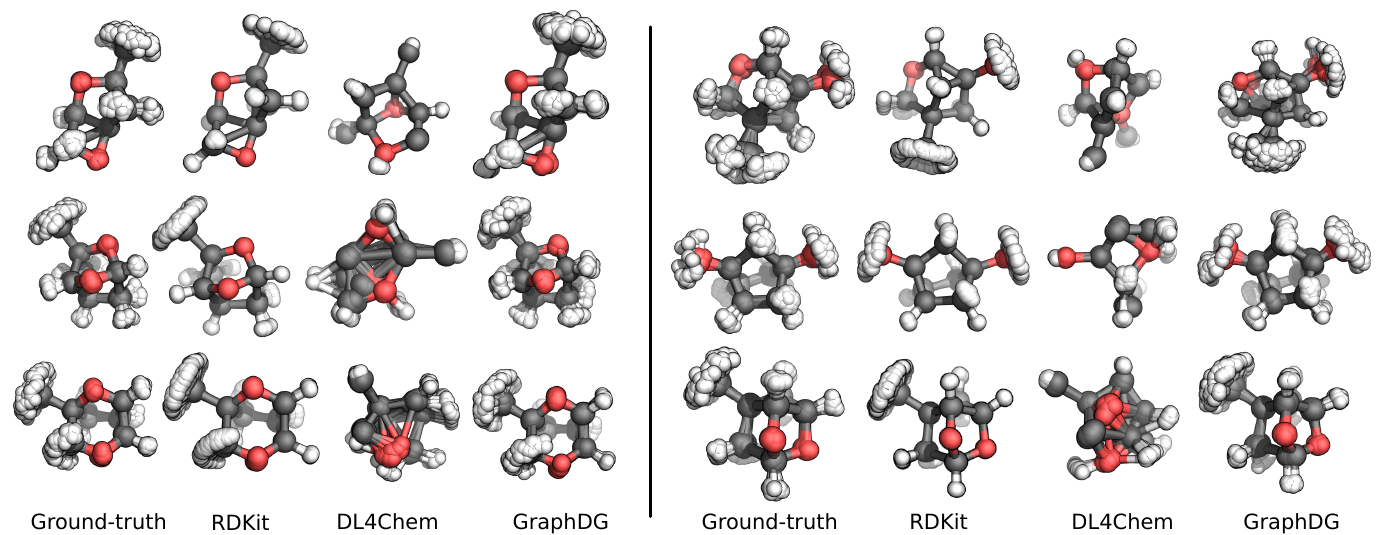}
\caption{
Overlay of 50 conformations from the ground-truth and three models based on six random molecular graphs from the test set.
C, O, and H atoms are colored gray, red, and white, respectively.
}
\label{fig:samples}
\end{figure*}
We passed the distances from our generative model to an EDG algorithm to obtain conformations.
For 99.9\% of the sets of distances, all triangle inequalities held.
For 83\% of the molecular graphs, the algorithm succeeded which is 7~pp higher than the success rate we observed for \textsc{RDKit}.
For each molecular graph in a test set, we generated 50 conformations with each method.
This took \textsc{DL4Chem}, \textsc{RDKit}, and \textsc{GraphDG} on average around hundreds of milliseconds per molecule.%
\footnote{All simulations were carried out on a computer equipped with an i7-3820 CPU and a GeForce GTX 1080 Ti GPU.}
In contrast, a single conformation in the \textsc{ISO17} dataset takes around a minute to compute.

To assess the approximations made in the IS scheme, we studied the overlap between $p(\mathbf{d} | z, \mathcal{G})$ for a given $\mathcal{G}$ and different samples of $z$.
We found experimentally that for 50 samples the overlap between the distributions is small.
This finding can be explained by the high dimensionality of $\mathbf{d}$ which is on average $\approx 60$.

In Fig.~\ref{fig:samples}, an overlay of these conformations of six molecules generated by the different methods is shown.
It can be seen that \textsc{RDKit}'s conformations show too little variance,
while \textsc{DL4Chem}'s structures are mostly invalid, which is due in part to its failure to predict the correct interatomic angles.
Our method slightly overestimates the structural variance (see, for example, Fig.~\ref{fig:samples}, top row, second column),
but produces conformations that are the closest to the ground-truth.

\subsection{Calculation of Molecular Properties}

We estimate expected molecular properties for molecular graphs from the test set with $N=50$ conformational samples each.
Due to their poor quality, we could not compute properties $\mathcal{O}(\mathbf{x})$, including the energy $E(\mathbf{x})$, for conformations generated with \textsc{DL4Chem},
and thus, this method is excluded from this analysis.
In Table~\ref{tab:props}, it can be seen that \textsc{RDKit} and \textsc{GraphDG} perform similarly well (computational details can be found in the Appendix).
However, both methods are still highly inaccurate for $E_\text{elec}$ (in practice, an accuracy of less than 5~kJ/mol is required).
Close inspection of the conformations shows that,
even though \textsc{GraphDG} predicts the most accurate distances overall,
the variances of certain strongly constrained distances (e.g., triple bonds) are overestimated so that the energies of the conformations increase drastically.

\begin{table}
\centering
\caption{Median difference in average properties between ground-truth and \textsc{RDKit} and \textsc{GraphDG}:
total electronic energy $E_\text{elec}$ (in kJ/mol),
the energy of the HOMO and the LUMO $\epsilon_\text{HOMO}$ and $\epsilon_\text{LUMO}$, respectively (in eV),
and the dipole moment $\mu$ (in debye).
Reported are the results for molecular graphs from the test set, averaged over three train-test splits.
Standard errors are given in brackets.
}
\label{tab:props}
\begin{tabular}{l|cc}
\toprule
                        &    \textsc{RDKit}     &   \textsc{GraphDG}  \\
\midrule
$E_\text{elec}$         &   $42.7 \; (4.3)$     &   $58.0 \; (21.0)$  \\
$\epsilon_\text{HOMO}$  &   $0.08 \; (0.04)$    &   $0.10 \; (0.05)$  \\
$\epsilon_\text{LUMO}$  &   $0.15 \; (0.03)$    &   $0.09 \; (0.05)$  \\
$\mu$                   &   $0.29 \; (0.05)$    &   $0.33 \; (0.09)$  \\
\bottomrule
\end{tabular}
\end{table}

\section{Limitations}

The first limitation of this work is that the CVAE can sample invalid sets of distances for which there exists no 3D structure.
Second, the \textsc{Conf17} benchmark covers only a small portion of chemical space.
Finally, a large set of auxiliary edges would be required to capture long-range correlations (e.g., in proteins).
Future work will address these points.

\section{Conclusions}

We presented \textsc{GraphDG}, a transferable, generative model that allows sampling from a distribution over molecular conformations.
We developed a principled learning representation of conformations that is based on distances between atoms.
Then, we proposed a challenging benchmark for comparing molecular conformation generators.
With this benchmark, we show experimentally that conformations generated by \textsc{GraphDG} are closer to the ground-truth than those generated by other methods.
Finally, we employ our model as a proposal distribution in an IS integration scheme to estimate molecular properties.
While orbital energies and the dipole moments were predicted well,
a larger and more diverse dataset will be necessary for meaningful estimates of electronic energies.
Further, methods have to be devised to estimate how many conformations need to be generated to ensure all important conformations have been sampled.
Finally, our model could be trained on conformational distributions at different temperatures in a transfer learning-type setting.

\section*{Acknowledgments}

We would like to thank the anonymous reviewers for their valuable feedback.
We further thank Robert Perharz and Hannes Harbrecht for useful discussions and feedback.
GNCS acknowledges funding through an Early Postdoc.Mobility fellowship by the Swiss National Science Foundation (P2EZP2\_181616).

\bibliography{references}
\bibliographystyle{icml2020}

\newpage
\clearpage
\appendix

\section{\textsc{Conf17} Benchmark}
\label{sec:conf17_appendix}

\subsection{Data Generation}

The \textsc{ISO17} dataset \citep{Schutt2017} was processed in the following way.
First, conformations in which could not be parsed by the tool \textsc{XYZ2Mol} \citep{JensenXYZ2mol} were discarded.
Second, the molecular graphs were augmented by adding auxiliary edges for reasons described in the main text.
This can lead to an over-specification of the system's geometry, however, this did not pose a problem in our experiments.

\subsection{Input Features}

Below we list the node and edge features in the \textsc{Conf17} benchmark.

\begin{table}[H]
\caption{Node features.}
\label{tab:node_feats}
\begin{center}
\begin{tabular}{lll}
\toprule
Feature & Data Type & Dimension \\
\midrule
atomic number & one-hot (H, He, ... F) & 9\\
chiral tag & one-hot (R, S, and None) & 3 \\
\bottomrule
\end{tabular}
\end{center}
\end{table}

\begin{table}[H]
\caption{Edge features.}
\label{tab:edge_feats}
\begin{center}
\begin{tabular}{lp{9em}l}
\toprule
Feature & Data Type & Dimension \\
\midrule
kind & one-hot ($E_\text{bond}$, $E_\text{angle}$, or $E_\text{dihedral}$) & 3 \\
stereo chemistry & one-hot (E, Z, Any, None) & 4 \\
type & one-hot (single, double, triple, aromatic or None) & 5 \\
is aromatic & binary & 1 \\
is conjugated & binary & 1 \\
is in ring of size & one-hot (3, 4, \ldots, 9) or None & 7 \\
\bottomrule
\end{tabular}
\end{center}
\end{table}

\section{Model Architecture}
\label{sec:arch}

The source code of the model (including pre-processing scripts) is available online \url{https://github.com/gncs/graphdg}.
In Table~\ref{tab:models}, the model architecture are summarized.
After each hidden layer, a ReLU non-linearity is used.
In Table~\ref{tab:hyper}, all hyperparameters are listed.

\begin{table}[ht]
\centering
\caption{Model architecture for generative model.}
\label{tab:models}
\begin{tabular}{@{}cr@{}}
\toprule
Network & \multicolumn{1}{c}{Output Layers}  \\
\midrule
$F_{\text{enc}, v}$ & $50, 50, 10$  \\
$F_{\text{enc}, e}$ & $50, 50, 10$  \\
$\text{MP}^{(t)}_\text{enc}$ & $50, 50, 10$  \\
$\text{MP}^{(t)}_\text{dec}$ & $50, 50, 10$  \\
$R_{\text{dec}, v}$ & $50, 50, 1$  \\
$R_{\text{dec}, e}$ & $50, 50, 1$  \\
\bottomrule
\end{tabular}
\end{table}

\begin{table}[ht]
\centering
\caption{Hyperparameters used throughout this work.}
\label{tab:hyper}
\begin{tabular}{@{}cc@{}}
\toprule
Hyperparameter & Value \\
\midrule
Minibatch size & 32 \\
Learning rate & 0.001 \\
Number of convolutions $T$ & 3 \\
\bottomrule
\end{tabular}
\end{table}

\section{Computational Details}
\label{sec:comp_appendix}

\subsection{Quantum-Chemical Calculations}

All quantum-chemical calculations were carried out with
the \texttt{PySCF} program package (version 1.5) \citep{Sun2018}
employing the exchange-correlation density functional PBE \citep{Perdew1996a},
and the def2-SVP \citep{Weigend2005,Weigend2006} basis set.

Conformations generated by \textsc{DL4Chem} did not succeed as some atoms were too close to each other.
Self-consistent field algorithms in quantum-chemical software such as \texttt{PySCF} do not converge for such molecular structures.

With quantum-chemical methods, we calculate several properties that concern the states of the electrons in the conformation.
These are the total electronic energy $E_\text{elec}$,
the energy of the electron in the highest occupied molecular orbital (HOMO in eV) $\epsilon_\text{HOMO}$,
the energy of the lowest unoccupied molecular orbital (LUMO in eV) $\epsilon_\text{LUMO}$,
and the norm of the dipole moment $\mu$ (in debye).

\subsection{Euclidean Distance Geometry}

We refer the reader to \citet{Havel2002} for theory on EDG, algorithms, and chemical applications.
In summary, the EDG procedure consists of the following three steps:
\begin{enumerate}
\item Bound smoothing: extrapolating a complete set of lower and upper limits on all the distances from the sparse set of lower and upper bounds.
\item Embedding: choosing a random distance matrix from within these limits, and computing coordinates that are a certain best-fit to the distances.
\item Optimization: optimizing these coordinates versus an error function which measures the total violation of the distance
(and chirality) constraints.
\end{enumerate}

We use the EDG implementation found in \textsc{\textsc{RDKit}} \citep{Riniker2015} with default settings.

\section{Generation of Conformations}
\label{sec:confs_appendix}

\begin{figure*}
\begin{center}
\includegraphics[width=\textwidth]{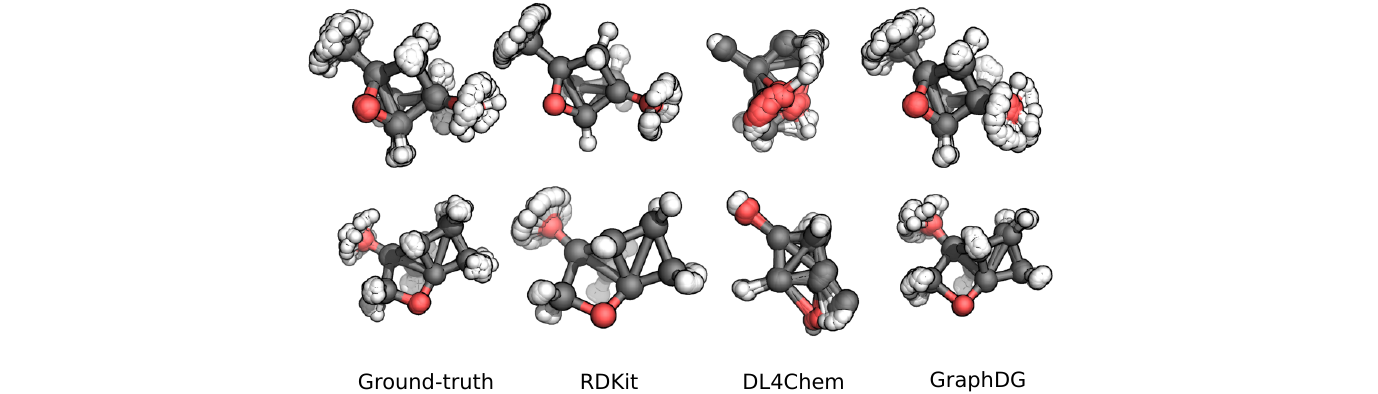}
\caption{
Overlay of 50 conformations from the ground-truth, \textsc{RDKit}, \textsc{DL4Chem}, and \textsc{GraphDG}
based on two random molecular graphs from a test set.
C, O, and H atoms are colored gray, red, and white, respectively.
}
\label{fig:samples_appendix}
\end{center}
\end{figure*}

Fig.~\ref{fig:samples_appendix} shows an overlay of 50 conformations from the ground-truth,
\textsc{RDKit}, \textsc{DL4Chem}, and \textsc{GraphDG}
based on two random molecular graphs from a test set.

\section{Distributions over Distances}
\label{sec:dists_appendix}

We show the marginal distributions
$p(d_{k} | G)$ and $p(d_{i}, d_{j} | G)$ of ground-truth and predicted distances (in \AA{})
for additional molecules from a test set.

\begin{figure*}[t]
\begin{center}
\includegraphics{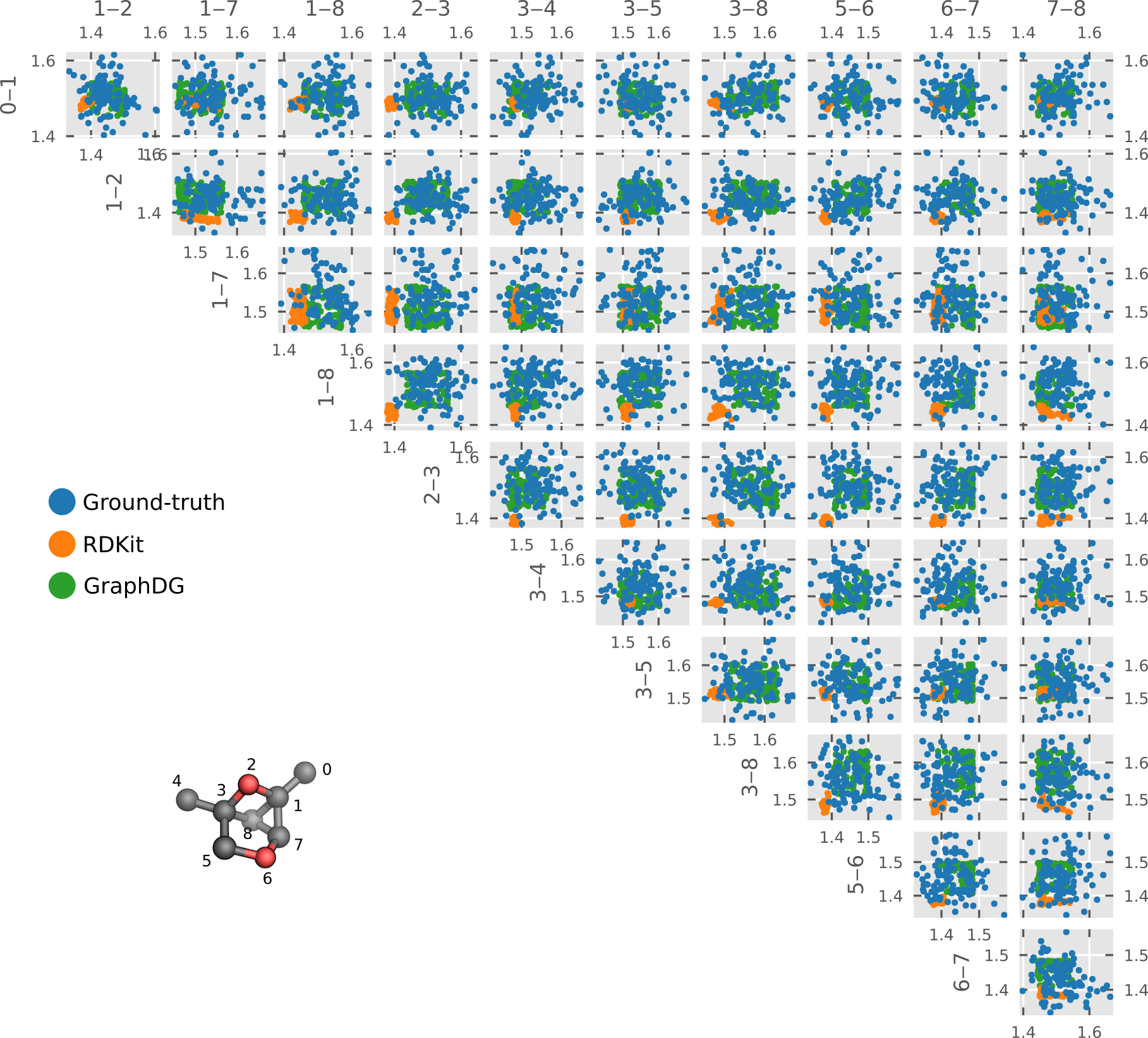}
\caption{
Marginal distributions $p(d_{i}, d_{j} | G)$ of ground-truth and predicted distances for a molecular graph
from a test set (in \AA{}).
Here, $d_{i}$ and $d_{j}$ are restricted to edges representing bonds between C and O atoms.
In the 3D structure of the molecule, carbon and oxygen atoms are colored gray and red, respectively.
H atoms are omitted for clarity.
}
\label{fig:scatter}
\end{center}
\end{figure*}

\begin{figure*}[t]
\begin{center}
\includegraphics{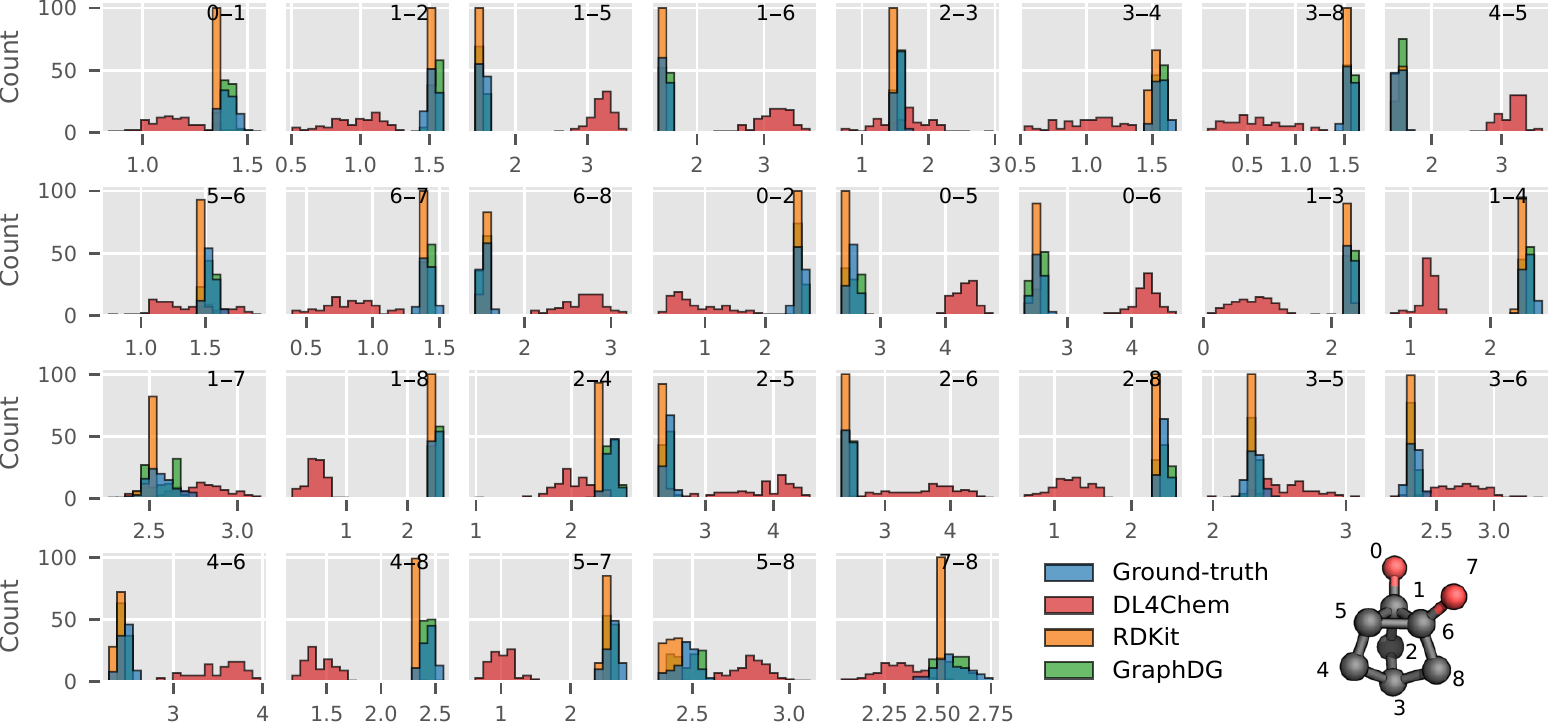}
\caption{
See caption of Fig.~\ref{fig:hists}
}
\end{center}
\end{figure*}

\begin{figure*}[t]
\begin{center}
\includegraphics{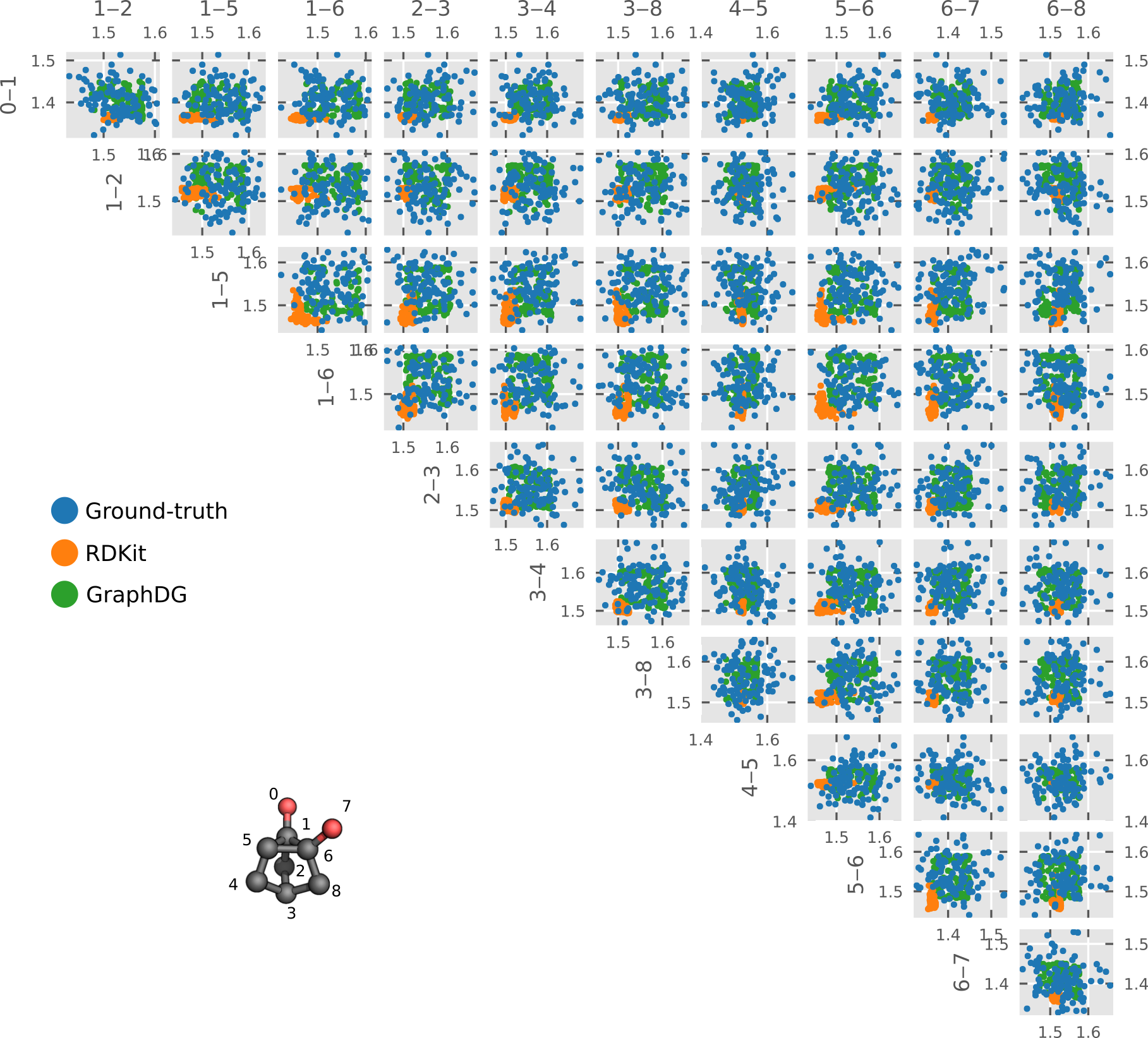}
\caption{
See caption of Fig.~\ref{fig:scatter}
}
\end{center}
\end{figure*}

\begin{figure*}[t]
\begin{center}
\includegraphics{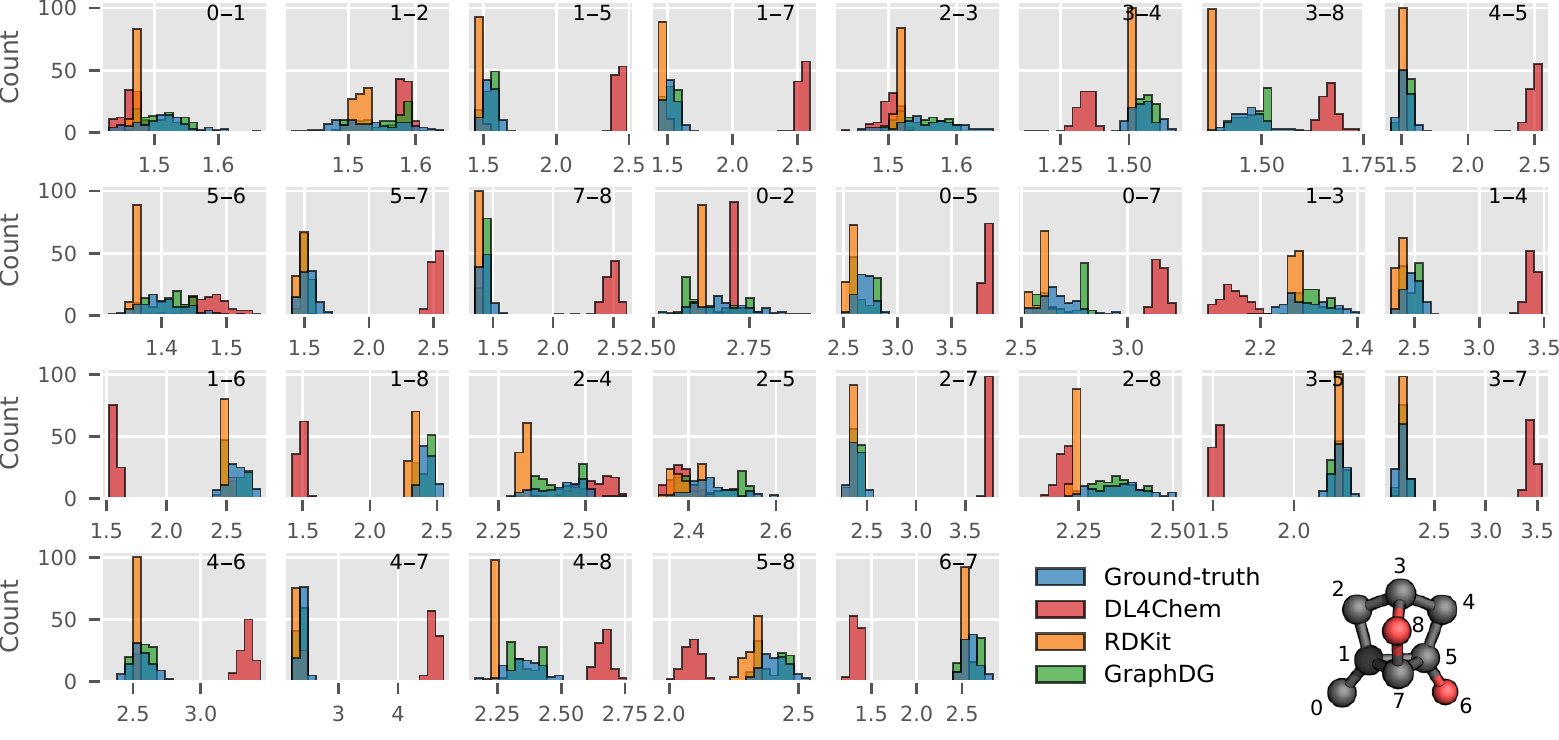}
\caption{
See caption of Fig.~\ref{fig:hists}
}
\end{center}
\end{figure*}

\begin{figure*}[t]
\begin{center}
\includegraphics{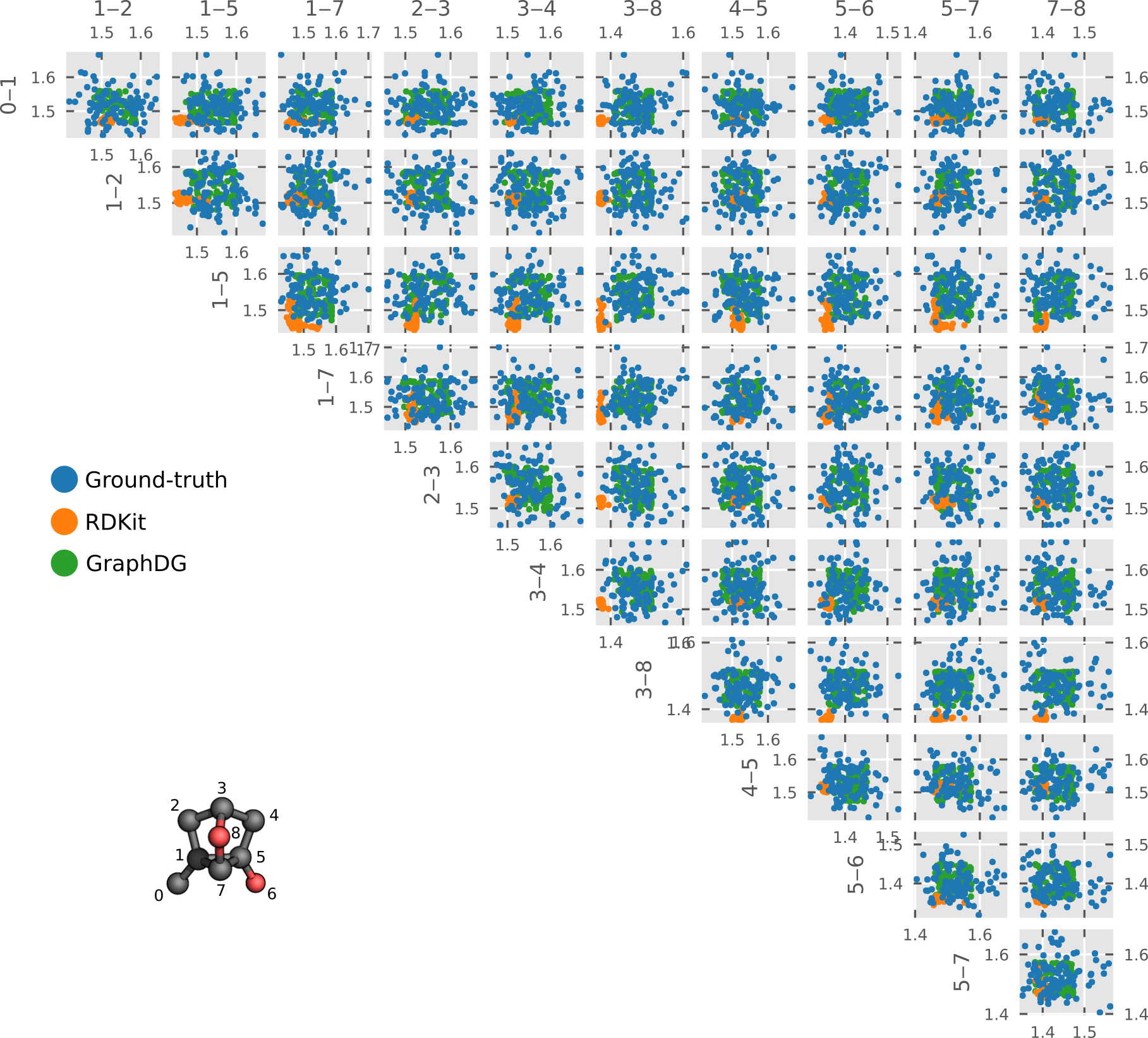}
\caption{
See caption of Fig.~\ref{fig:scatter}
}
\end{center}
\end{figure*}

\begin{figure*}[t]
\begin{center}
\includegraphics{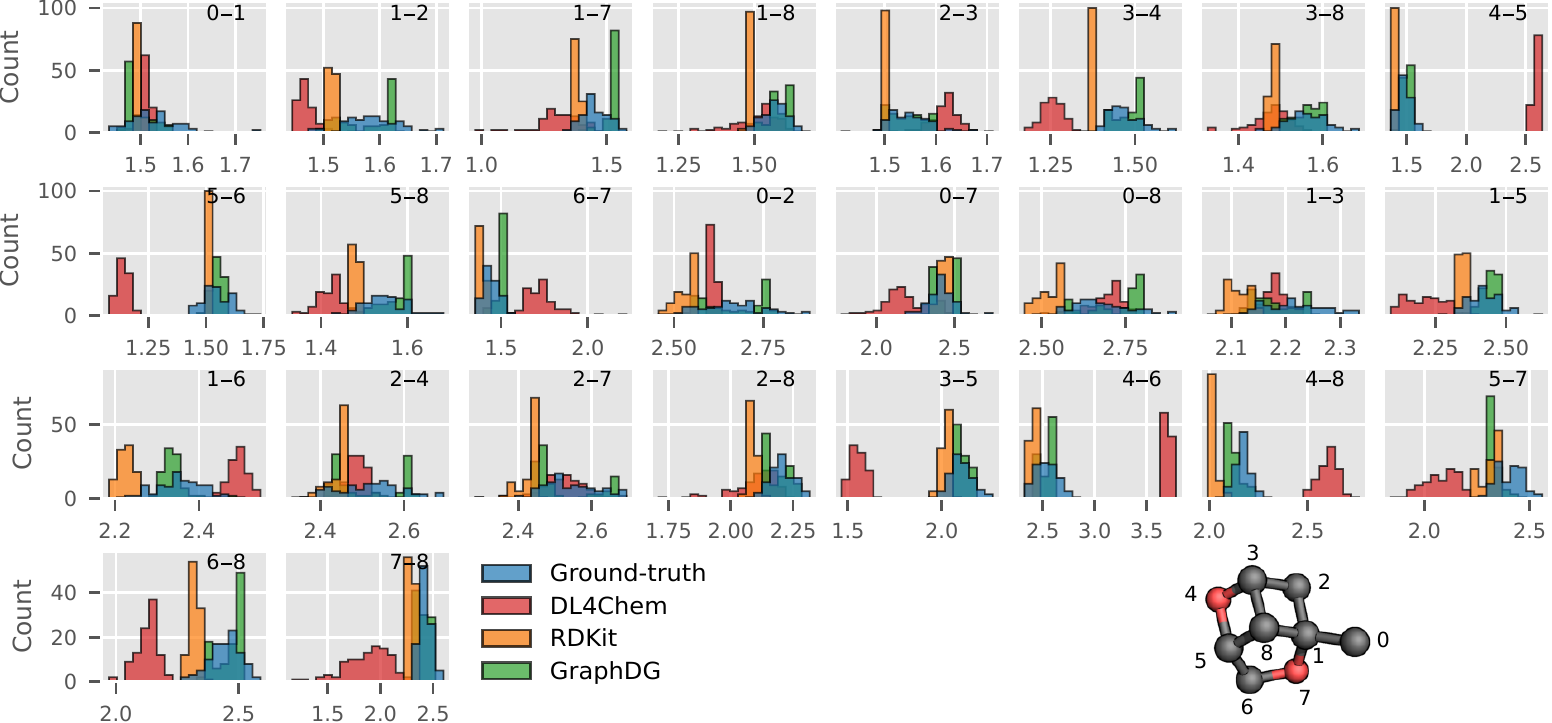}
\caption{
See caption of Fig.~\ref{fig:hists}
}
\end{center}
\end{figure*}

\begin{figure*}[t]
\begin{center}
\includegraphics{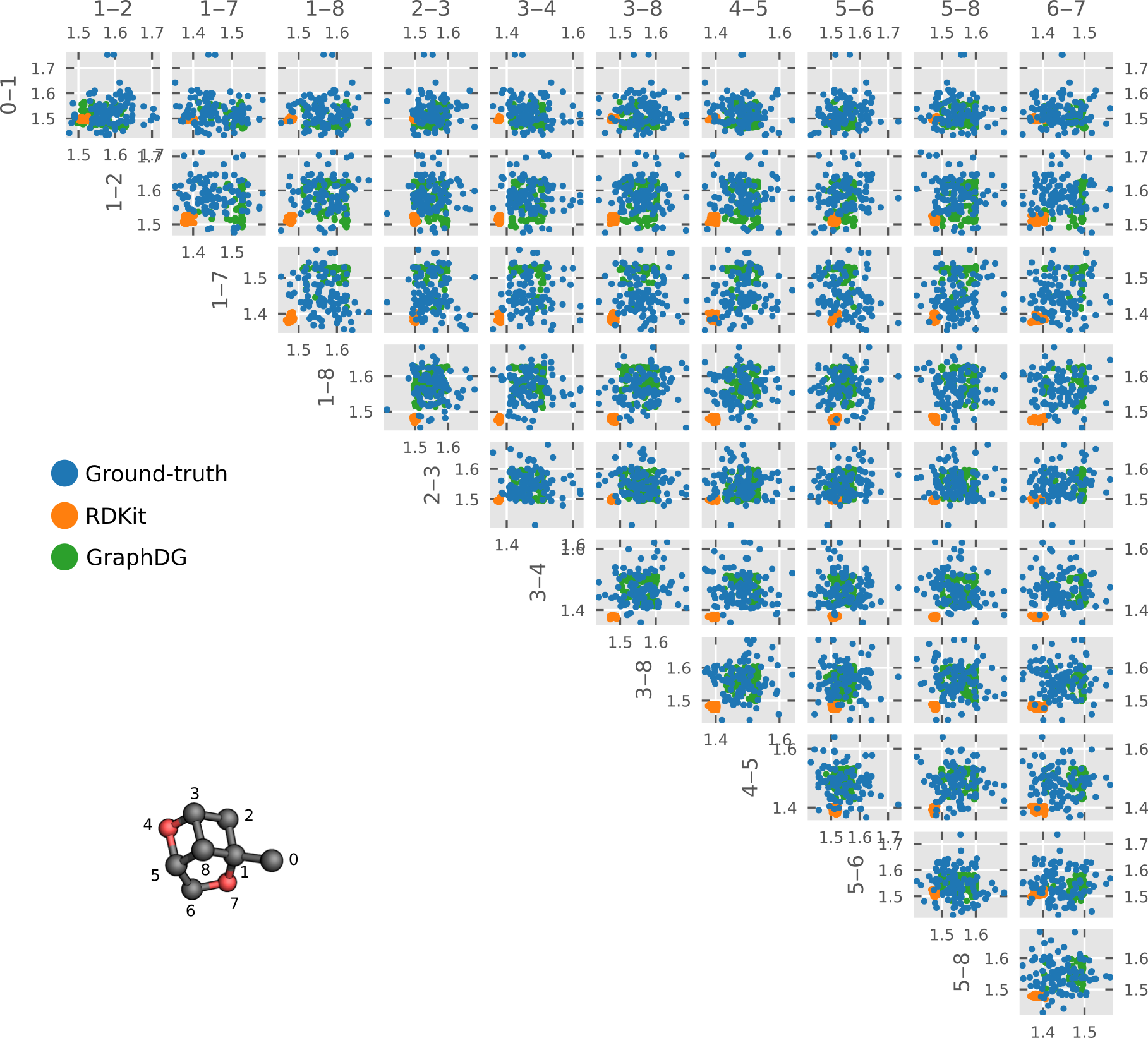}
\caption{
See caption of Fig.~\ref{fig:scatter}
}
\end{center}
\end{figure*}

\end{document}